\documentclass[conference,10pt,letterpaper]{IEEEtran}
\usepackage[mathscr]{eucal}
\usepackage{ifpdf}
\usepackage{cite}
\usepackage{graphicx}
\usepackage[cmex10]{amsmath}
\usepackage{amssymb}
\usepackage{algorithmic}
\usepackage{units}
\usepackage{setspace}
\usepackage{algorithm}
\usepackage{array}
\usepackage[tight,footnotesize]{subfigure}
\usepackage{amsthm}
\usepackage{multirow}
\usepackage{enumerate}
\usepackage{color}
\usepackage{mathtools}
\usepackage{subfigure}
\usepackage{tabularx}

\newcounter{assume}

\newtheorem{theorem}{Theorem}

\newcommand{\resl}[1]{}
\newcommand{\tosl}[1]{}
\newcommand{\cosl}[1]{}

\begin{document}
\title{Multi-armed Bandits with Application to \\ 5G Small Cells}
\author{
\IEEEauthorblockN{Setareh Maghsudi and Ekram Hossain, \textit{Fellow, IEEE}}
}
\maketitle
\begin{abstract}
Due to the pervasive demand for mobile services, next generation wireless networks 
are expected to be able to deliver high date rates while wireless resources become 
more and more scarce. This requires the next generation wireless networks to move 
towards new networking paradigms that are able to efficiently support resource-demanding 
applications such as personalized mobile services. Examples of such paradigms foreseen 
for the emerging fifth generation (5G) cellular networks include very densely deployed 
small cells and device-to-device communications. For 5G networks, it will be imperative to 
search for spectrum and energy-efficient solutions to the resource allocation problems 
that i) are amenable to distributed implementation, ii) are capable of dealing with 
uncertainty and lack of information, and iii) can cope with users' selfishness. The 
core objective of this article is to investigate and to establish the potential of 
multi-armed bandit (MAB) framework to address this challenge. In particular, we 
provide a brief tutorial on bandit problems, including different variations and 
solution approaches. Furthermore, we discuss recent applications as well as future 
research directions. In addition, we provide a detailed example of using an MAB model 
for energy-efficient small cell planning in 5G networks.

\let\thefootnote\relax\footnotetext{The authors are with the Department of Electrical 
and Computer Engineering, University of Manitoba, Winnipeg, MB, Canada (e-mail: 
maghsudi.setareh@gmail.com, ekram.hossain@umanitoba.ca).}
\end{abstract}

{\em Keywords}:- 5G cellular networks, small cells, single-agent and multi-agent multi-armed 
bandit (MAB) problems, correlated equilibrium, Nash equilibrium.
%


\section{Introduction}
\label{Sec:Intro}
Due to the ever-increasing need for mobile services, we expect a massive 
growth in demand for wireless services in the years to come. As a result, 
future mobile networks are expected to accommodate new communications, 
networking, and energy-efficiency concepts, for instance, small cells, device-to-device 
(D2D) communications, and energy harvesting. In order to realize 
such novel concepts, system designers face some fundamental challenges. For 
instance, while in traditional cellular networks some channel state information 
(CSI) is assumed to be available at some base station (BS) and wireless devices, 
such information is not likely to be affordable in emerging hierarchical or 
distributed networks. Centralized 
resource management in such networks (especially in a very dense deployment scenario) will incur excessive complexity, and therefore,
distributed approaches will need to be developed, which might trigger competitive 
user behavior. Moreover, while any planning and scheduling in distributed 
networks depends heavily on the available energy at network nodes, energy 
levels are not observable in general. Therefore, it is
evident that in future, efficient and robust wireless communications design 
needs to deal with inherent uncertainty and lack of information, as well as 
possible competition for resources, in a distributed manner. As a result, it 
becomes imperative to search for new mathematical tools to deal with networking 
problems, which in general involve both uncertainty and conflict. 

Multi-armed bandit (MAB) is a class of sequential optimization problems, 
where, in the most seminal form, given a set of arms (actions), a player 
pulls an arm at each round in order to receive some reward. The rewards 
are not known to the player in advance; however, upon pulling any arm, 
the instantaneous reward of that arm is revealed. In such an unknown setting, 
at each trial, the player may lose some reward (or incur some cost) due 
to not selecting the best arm instead of the played arm. This loss is 
referred to as \textit{regret}. The player decides which arm to pull in 
a sequence of trials so that its accumulated regret over the horizon is 
minimized, or its discounted reward over the horizon is maximized.

MAB modeling brings a variety of advantages to the research in the area 
of wireless communications and networking; some of them are briefly described 
in the following.
\begin{itemize}
\item As mentioned before, every MAB model includes uncertainty, resulted 
      from lack of prior knowledge as well as strictly limited feedback. 
      This corresponds to a primal and natural feature of future distributed 
      wireless networks, where providing nodes with information yields 
      excessive feedback and overhead cost.
\item In recent years, multi-agent MAB has attracted great attention, which 
      allows to introduce the concept of conflict to the seminal single-agent 
      MAB. Multi-agent MAB models can be beneficially applied to solve wireless 
      networking problems such as distributed 
      resource management or interference coordination problems in a distributed manner.
\item MAB benefits from many variations in the model and therefore is 
      able to accommodate many distinct conditions in wireless networks, 
      for instance different levels of information availability and 
      different types of randomness.      
\end{itemize}
In this article, we provide a brief tutorial overview of  both single-agent 
and multi-agent MABs, including different variations in the model. Furthermore, we 
study state-of-the-art of applications of MAB models in wireless resource 
allocation, and discuss future research directions and open problems. Finally, 
we describe a new bandit-theoretical model for energy-efficient small cells 
planning in 5G networks.


\section{Single-Agent Multi-armed Bandit}
\label{sec:SPMAB}
Single-agent multi-armed bandit (SA-MAB) problem was first introduced in \cite{Robbins52:SASDE}. 
In the most seminal setting, the problem models an agent facing the challenge of sequentially 
selecting an arm from a set of arms in order to receive an {\em a priori} unknown reward, drawn from 
some unknown reward generating process. As a result of lack of prior information, at each trial, 
the player may choose some inferior arm in terms of reward, yielding some \textit{regret} that 
is quantified by the difference between the reward that would have been achieved had the player 
selected the best arm and the actual achieved reward. In such an unknown setting, the player 
decides which arm to pull in a sequence of trials so that its accumulated regret over the game 
horizon is minimized. This problem is an instance of \textit{exploration-exploitation} dilemma, 
i.e., the trade-off between taking actions that yield immediate large rewards on the one hand 
and taking actions that might result in larger reward only in future, for instance activating 
an inferior arm only to acquire information, on the other hand. Therefore, SA-MAB is mainly 
concerned with a sequential online decision making problem in an unknown environment. A solution 
of a bandit problem is thus a decision making strategy called \textit{policy} or \textit{allocation 
rule}, which determines which arm should be played at successive rounds. Policies are in general 
evaluated based on their regret or discounted reward performance. 

As mentioned before, MAB benefits from a wide range of variations in the setting. In the event that 
arms have different states, the reward depends on arms' states, and the bandit model is referred to 
as \textit{stateful (Markovian)}. Otherwise, the model is \textit{stateless}, which itself can be 
divided to few subsets. As stateful and stateless bandits are inherently different, we discuss them 
separately in the following.  
\subsection{Stateful (Markovian) Bandits}
\label{subsec:stateful}
In a stateful bandit model, every arm is associated with some finite 
state space. Upon being pulled, each arm pays some positive reward that 
is drawn from some stationary distribution associated to the current 
state of that arm. Arms' states change over time according to some 
stochastic model, which is considered to be a Markov process; that is, 
it satisfies the Markov property. Roughly speaking, a process satisfies 
the Markov property if its future depends solely on its present state 
rather than its full history. This type of MAB is also referred to as 
\textit{Markovian bandit}. Note that in this formulation, after each 
round, the reward as well as the state of only the played arm is revealed, 
and the mechanism of state transition is unknown. A Markovian MAB model 
can be thus defined formally as $\mathfrak{B}\coloneqq \left\{\mathcal{M},
\mathcal{S}_{m},\pi_{m,s},\mathbf{T}_{m},\mu_{m}\right\}$, where $m \in 
\mathcal{M}$, and 
\begin{itemize}
\item $\mathcal{M}$ is the set of arms,
\item $\mathcal{S}_{m}$ is the finite set of states of arm $m$,
\item $\pi_{m,s}$ is the stationary reward distribution of arm 
      $m$ in state $s \in \mathcal{S}_{m}$ with mean $\mu_{m,s}$, 
\item $\mathbf{T}_{m}$ is the state transition matrix of arm $m$,
\item $\mu_{m}$ is the average reward of arm $m$. 
\end{itemize}

For Markovian bandits, the mean reward of arm $m$, $\mu_{m}$, is its 
expected reward under the rewards' stationary distribution in different 
states. That is, $\mu_{m}=\sum_{s \in \mathcal{S}_{m}} \mu_{m,s}\pi_{m,s}$. 
Define $\mu^{*} \coloneqq \underset{m \in \mathcal{M}}{\max}~~\mu_{m}$. 
Also, let $s_{m,t}$ denote the state of some arm $m$ at time $t$. The 
selected arm at each time $t$ is denoted by $I_{t}$. Then, for Markovian 
bandit, the cumulative regret up to time $n$, denoted by $R_{\textup{Mar}
,n}$, is defined as
\begin{equation}
\label{eq:Markovian}
R_{\textup{Mar},n}=n\mu^{*}-\mathbf{E}\left[\sum_{t=1}^{n}\mu_{I_{t},s_{I_{t},t}} \right],
\end{equation}
where the expectation, $\mathbf{E}\left[ \cdot \right]$, is taken with 
respect to the random draw of both states and the random actions of the 
policy. The general problem is to find the optimal policy, where optimality 
is defined in the sense of minimum regret.

Stateful bandit models are conventionally divided into the two following 
groups: i) rested (frozen) bandits, where at each round, only the state 
of the played arm changes, and ii) restless bandits, where the state of 
every arm is subject to change after each round of play.

Markovian bandit problems are often solved using \textit{indexing policies}. 
Roughly speaking, at each round, a real scalar value, referred to as index, 
is associated to each arm. The index of any arm is counted as a measure of 
the reward that can be achieved by activating that arm in the current state. 
The arm with the highest index is played at each round. The optimal indexing 
policy is not unique for different bandit models; it depends on a variety of 
factors such as reward generating processes and arms being restless or frozen. 
An example of an indexing policy can be found in \cite{Liu13:LCW}.

Note that if the arms' states are entirely observable and the state transition 
matrices are known, then the criterion for action selection is conventionally to 
maximize the discounted reward, instead of minimizing the regret. Such problems 
belong to bandit models, and are solved by using indexing policies as well, with 
an example being Gittins index \cite{Gittins11:MABAI}. It should be however noted 
that although the Gittins index has been calculated in closed form for a variety 
of stochastic reward processes, before using any indexing policy, the indexability 
of the problem under investigation has to be verified, which is in general not 
trivial. In addition to the Gittins indices, Whittle's indexing policy \cite{Gittins11:MABAI} 
is well-known and often used to solve restless bandits. This solution however does 
not hold in general, and restless bandit problems are intractable in many cases. 
Important approximation algorithms for some families of restless 
bandit problems can be found in \cite{Guha10:AARB} and \cite{Bertsimas00:RBLP}.
\subsection{Stateless Bandits}
\label{subsec:stateless}
In stateless bandit model, arms do not have any specific state. The arms' reward generating 
processes, however, are not necessarily stationary. A stateless MAB model is thus formally 
defined as $\mathfrak{B}\coloneqq \left\{\mathcal{M},\mu_{m,t}\right\}$, where $m \in \mathcal{M}$, 
and
\begin{itemize}
\item $\mathcal{M}$ is the set of arms, and
\item $\mu_{m,t}$ is the average reward of arm $m$ at time $t$. 
\end{itemize}
Based on the nature of reward generating process, stateless MABs can be divided into few 
categories. Before proceeding to describe these categories, however, we provide a notion 
of regret that is widely used to analyze almost every stateless bandit problem, regardless 
of the category to which it belongs. Let $g_{m,t}$ and $I_{t}$ denote, respectively the instantaneous 
reward of arm $m$ and the selected arm, both at time $t$. Then, the regret of any policy 
for stateless bandits up to time $n$, referred to as \textit{external regret}, which is denoted 
by $R_{\textup{Ext},n}$, is defined as
\begin{equation}
\label{eq:External}
R_{\textup{Ext},n}=\max_{m \in \mathcal{M}}\mathbf{E}\left [\sum_{t=1}^{n}g_{m,t}\right]
-\mathbf{E}\left [\sum_{t=1}^{n}g_{I_{t},t}\right].
\end{equation}
Given the definition of external regret, the optimal solution of a stateless bandit problem 
is a policy that minimizes the external regret. Now we are in a position to study two important 
branches of stateless bandit problems, namely, stochastic and adversarial bandits. 
\subsubsection{Stochastic Bandits}
\label{subsec:stochastic}
Stochastic MAB is a stateless bandit model, where arms are not associated with states, and 
the reward generating processes are stochastic. In other words, the series of rewards of each 
arm are state-independent, drawn from some specific density function, which can be stationary 
or non-stationary. For the stationary case, $\mu_{m,t}=\mu_{m}$ for all $m \in \mathcal{M}$. 
In this case, the first term on the right hand side of (\ref{eq:External}) yields $n\mu^{*}$ 
where $\mu^{*} \coloneqq \underset{m \in \mathcal{M}}{\max}~~\mu_{m}$. For the non-stationary 
case, the first term yields $\sum_{t=1}^{n}\mu^{*}_{t}$, with $\mu^{*}_{t} \coloneqq \underset{m 
\in \mathcal{M}}{\max}~~\mu_{m,t}$. 

In order to solve the stochastic bandit problem, many methods have been developed so far that 
are based on the \textit{upper confidence bound (UCB)} policy \cite{Bubeck12:RASNS}. The 
basic idea to deal with the exploration-exploitation dilemma  here is to estimate an upper 
bound of the mean reward of each arm at some fixed confidence level. The arm with the 
highest estimated bound is then played. Such methods, however, are only applicable when 
the rewards of each arm are independent and identically distributed, or in other words, 
the arm is stationary. In order to deal with non-stationarity, algorithms mostly use some 
statistical test in order to detect changes in the distribution. To elaborate, consider 
a sequence of independent random variables with some density that depends on some scalar 
parameter $\theta$. The initial value of this parameter is $\theta_{0}$, and after an 
unknown point of time, its value changes to $\theta_{1}$. \textit{Change-point detection} 
is performed by using some statistical test, for example, \textit{generalized likelihood 
ratio} or \textit{Page-Hinkly} test, which not only identifies the changes, but also estimates 
the change time. Note that in general it is assumed that not too many of such change points 
exist. There are also some adapted versions of the seminal UCB algorithm that detect changes in 
the sample mean reward of arms in order to deal with time-variant statistical characteristics. 
For example, in \textit{discounted UCB}, the rewards are weighted so that recent outcomes are 
emphasized when calculating the sample mean rewards, based on which the next arm is selected. 
Or, in \textit{sliding-window UCB}, the sample mean reward of each arm is calculated only 
over a fixed-length window of recent rewards, and not by using the entire reward history.
\subsubsection{Adversarial (Non-stochastic) Bandits}
\label{subsec:adversarial}
Adversarial bandit models are similar to stochastic ones in being stateless, with the 
difference that the series of rewards of each arm cannot be attributed to any specific 
distribution; in other words, reward are non-stochastic. Generally, in an adversarial 
setting, mixed strategies are used to select an arm. That is, at each time $t$, the 
agent selects a probability distribution $\mathbf{P}_{t}=\left(p_{1,t},...,p_{m,t},...,
p_{M,t}\right)$ over arms, and plays arm $m$ with probability $p_{m,t}$. In such a setting, 
\textit{in addition} to external regret given by (\ref{eq:External}), we define another 
important notion of regret, namely \textit{internal regret}, as follows:
\begin{equation}
\label{eq:Internal}
R_{\textup{Int},n}=\max_{m,l \in \mathcal{M}}R_{\left(m \to l \right),n}=
\max_{m,l \in \mathcal{M}}\sum_{t=1}^{n}p_{m,t}\left(g_{l,t}-g_{m,t}\right).
\end{equation}
By definition, external regret compares the expected reward of the current mixed strategy 
with that of the best fixed action in the hindsight, but fails to compare the rewards 
achieved by changing actions in a pairwise manner. Internal regret, in contrast, compares 
actions in pairs. Later, we shall see that the notion of internal regret is in close relation 
with correlated equilibria in games.

Adversarial bandits are often solved by using \textit{potential-based} or \textit{weighted 
average} algorithms \cite{Bubeck12:RASNS}. In the most seminal form of this approach, at each 
trial, a mixed strategy is calculated over the set of actions. The selection probability of 
each action is proportional to its average regret performance in the past, possibly weighted 
by a specific potential function (for example the exponential function). Accordingly, the 
actions with better past performance (lower average regret) are more likely to become activated 
in the future, and \textit{vice versa}. 
\subsection{Other Important Bandit Models}
\label{subsec:other}
\subsubsection{Contextual (Covariate) Bandits}
\label{subsec:contextual}
In the basic bandit model, at each round, the agent only observes the reward of the played 
actions, and hence has to select its future actions only based on its past performance. In 
\textit{contextual bandits} (also called \textit{bandits with side information} or \textit{covariate 
bandits}), in contrast, at each round of decision making, some side information is revealed 
to the learning agent. The agent therefore learns the best mapping of contexts to arms. Note 
that this type of bandit models is distinguished only based on information availability, and 
the reward process can be still Markovian, stochastic or adversarial. In any case, aforementioned 
algorithms are adapted to solve contextual bandit problems as well. 
\subsubsection{Mortal Bandits}
\label{subsec:mortal}
While in the basic bandit model it is assumed that all arms are infinitely available, there 
are also some variants that distinguish the problems where this assumption does not hold. 
For instance, \textit{mortal bandits} assume that arms are available only for a finite time. 
The availability time can be either deterministic or stochastic, known or unknown. An example 
of solution approaches can be found in \cite{Chakrabarti08:MMAB}. The algorithm is called 
\textit{stochastic with early stopping}, and the main idea behind it is to abandon an arm 
as soon as realizing that its maximum possible future reward may not justify its retention. 
The analysis of such schemes is inherently different from the general bandit model, since in 
such problems the achievable accumulated reward from each arm is bounded even over infinite 
horizon.
\subsubsection{Sleeping Bandits}
\label{subsec:sleeping}
\textit{Sleeping bandits} refers to bandit problems where action set is time-varying. Clearly, 
in such models, not only the reward process, but also the arm availability at each round can 
be Markovian, adversarial or stochastic. Some solution approaches for different types of this 
problem can be found in \cite{Klein08:RB}. The core idea of these algorithms is to change the 
notion of regret from the basic one we have defined before. While the basic notion is based on 
the performance loss with respect to the best action in the hindsight, here the best \textit{ordering} 
of actions serves as the benchmark, as the best action might not be available in some rounds.
\begin{figure*}[t]
\centering
\includegraphics[width=0.55\textwidth]{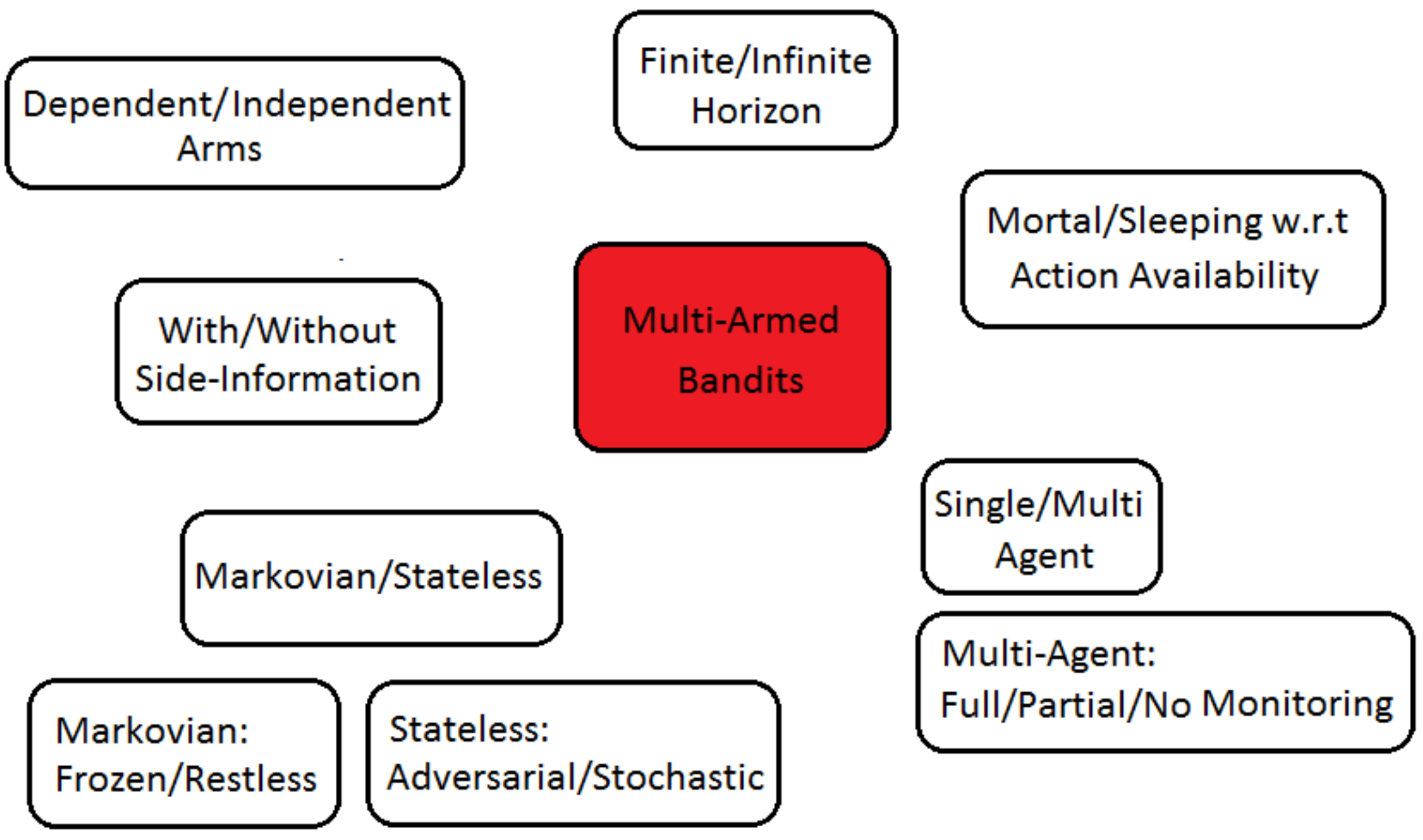}
\caption{Various types of MABs.}
\label{fig:Types}
\end{figure*}


\section{Multi-Agent (Game-Theoretical) Multi-armed Bandits}
\label{sec:MPMAB}
In multi-agent multi-armed bandits (MA-MAB), each player $k\in \mathcal{K}$ is assigned 
an action set $\mathcal{M}_{k} \subseteq \mathcal{M}$. Similar to the single-agent model, 
each agent selects an action in successive trials to receive an initially unknown reward; 
if multiple agents select an arm, the achieved reward is shared in an arbitrary manner. 
Therefore, in multi-agent setting, the reward of each agent depends on the joint action 
profile of all agents. Note that the action set, the played action and the reward achieved 
by each agent can be regarded as either private or public information, based on the specific 
system model and problem formulation. From the view point of each agent $k$, an MA-MAB 
model can be seen as a game with two players: the first player is agent $k$ itself, and 
the second player is the set of all other $K-1$ agents, whose joint action profile affects 
the reward achieved by agent $k$. This game might belong to any conventional category of 
games; for instance, it can be a potential or zero-sum game. 

Given no prior information, every agent requires to interact with the random environment 
in order to solve the problems that arise in an unknown reactive model, for instance long-term 
accumulated reward maximization, or average regret minimization. As a result of competition, 
which is inherent in non-cooperative multi-agent systems, single-agent bandit models that 
ignore the possible conflicts among multiple agents do not yield satisfactory outcomes; in 
particular, equilibrium is not guaranteed to be achieved. This is when game theory and multi-armed 
bandits meet and complement each other. Thereby, equilibrium arises as an asymptotic outcome 
of repeated interactions in an unknown environment among learning agents with bounded rationality 
that are provided with strictly limited information and aim at achieving long-term optimality 
in some sense. The aforementioned problem, i.e., efficient and convergent learning in a multi-agent 
setting, is currently under intensive investigation, mainly in computer science and mathematics. 
In the theory of wireless communications, there is also some ongoing research on game-theoretical 
bandits. In the following, we describe some important results in this area, with an emphasis 
on convergence to equilibrium.
\subsection{Convergence to Correlated Equilibrium}
\label{subsec:corrEq}
Recall the definition of internal regret given by (\ref{eq:Internal}) in Section 
(\ref{subsec:adversarial}). The following (simplified) theorem describes the 
relation between internal regret and the concept of correlated equilibrium in 
games.\footnote{The definition of correlated and Nash equilibrium are standard 
and therefore omitted here for the brevity of the article.} 
\begin{theorem}[\cite{Bianchi06:PLG}]
\label{th:CorrConvergenceO}
Consider a multi-agent bandit game with a set of players $\mathcal{K}$, and let 
$R_{\textup{Int},n}^{(k)}$ denote the internal regret of player $k \in \mathcal{K}$ 
at round $n$. If all agents play according to some policy that exhibits per-round 
vanishing internal regret (i.e., $\lim_{n\to \infty}\frac{1}{n}R_{\textup{Int},n}^{(k)}=0$), 
then the game converges to the set of correlated equilibrium in a time-average 
sense.
\end{theorem}
This concept is used for instance in \cite{Maghsudi13:JCSPC} to develop joint 
power control and channel selection strategies in distributed D2D networks.

Besides internal regret, there is another concept, namely \textit{calibrated forecasting}, 
which is related to correlated equilibria in games. Before explaining this relation, we 
briefly describe calibrated forecasting in the following. Consider a random process with 
a set $\mathcal{D}$ of $D$ outcomes. A forecaster predicts the outcome of the random process 
sequentially, and is referred to as \textit{calibrated} if, in the limit, the predicted 
outcome is equal to the true one. Now, consider a game with $K$ players, where each player 
$k$ selects its action from action set $\mathcal{M}_{k}$ (pure strategies). Then, for any 
player $k$, the joint action profile of its opponent is a random process with its set of 
outcomes being $\bigotimes_{i=1,i\neq k}^{K} \mathcal{M}_{k}$, where $\bigotimes$ denotes 
the Cartesian product. In such games, each player might apply a forecaster in order to predict 
the next joint action profile of its opponents, and uses that prediction to move strategically, 
for instance, by playing the best response to the predicted joint action profile. Note, however, 
such prediction naturally involves learning, and therefore, requires the knowledge of past 
actions of opponents; in other words, players should be able to observe the actions of each 
other. In the following theorem, we describe the relation between calibrated forecasting and 
correlated equilibria.
\begin{theorem}[\cite{Bianchi06:PLG}]
\label{th:CorrConvergenceT}
In any game, if every player plays by best responding to a calibrated forecast of 
its opponents' joint action profile, then the game converges to the set of correlated 
equilibrium in a time-average sense. 
\end{theorem}
Note that Theorem \ref{th:CorrConvergenceT} declares the convergence for general 
full-information games, where the game matrix is known to all players {\em a priori}. 
The theorem is generalized in \cite{Maghsudi13:CSNAD} to bandit games, and is 
applied as a basis to develop a convergent solution approach for non-stationary 
stochastic bandit models. The core idea is to combine calibrated forecasting with 
non-parametric regression, so that after some time, each player has some accurate 
estimate of the reward process of actions as a function of joint action profile 
and the next move of opponents.
\subsection{Convergence to Nash Equilibrium}
\label{subsec:nashEq}
While they are only few solution approaches for multi-agent bandit games that guarantee 
a convergence to correlated equilibria, converging to Nash equilibria seems to be an even 
more challenging task. There are some convergent approaches for special game classes such 
as potential games. For instance, in \cite{Chapman11:CLAUR}, algorithms are proposed for 
Nash convergence in potential games and games with more general forms of acyclicity. The 
basic idea therein is to combine Q-learning with stochastic better-reply or stochastic 
adaptive-play dynamics in order to develop a multi-agent version of Q-learning, which 
estimates the reward functions using novel forms of the $\epsilon$-greedy learning policy. 
Details can be found in \cite{Chapman11:CLAUR}. While multi-agent Q-learning converges only 
for some specific game classes, there are also 
some algorithms that converge in general games. A concept based on which multiple convergent 
bandit algorithms are proposed for general games is \textit{regret testing} \cite{Bianchi06:PLG}. 
Here the basic idea is to perform some sort of exhaustive search, in the sense that each player 
first selects a mixed strategy according to some predefined protocol, and then checks whether 
i) the incurred regret is less than a specific threshold; and ii) the selected mixed strategy 
is an approximate best response to the others' mixed strategies. Clearly, none of these can be 
concluded in one round of play; but, if the same mixed strategy is played during sufficiently 
many rounds, then each player can simply test the corresponding hypothesis. Different variants 
of this concept exhibit different convergence characteristics, for instance, convergence to Nash 
equilibrium in a time-average sense or convergence to the set of $\epsilon$-Nash equilibria, 
both for every generic game. 

In the area of wireless communications, most works consider a specific game (defined by 
a specific utility function) and propose a game-theoretical learning algorithm, whose 
convergence is established by some analysis based on the defined utility function. Such 
solutions, however, suffer from limited applicability, as they are tailored to converge 
only for the predefined games.


\section{State-of-the-Art and Future Research Directions}
\label{Sec:App}
In what follows, we first briefly review the current applications of MABs (in both 
single-agent and multi-agent settings). Afterwards, we discuss some open problems 
and potential research directions. Note that our focus in this article is on the 
applications of MAB models in solving resource allocation problems.
\subsection{State-of-the-Art}
\label{sec:current}
\subsubsection{Distributed Channel Selection}
\label{subsec:channelSelection}
Distributed channel selection is a widely-considered application of MABs. When cast 
as a MAB, channels are considered as arms, and reward processes are some functions 
of signal-to-noise ratio (SNR) or signal-to-interference-plus-noise ratio (SINR). 
Reward functions might evolve as Markovian, stochastic or adversarial, and any multiple 
access protocol, orthogonal and non-orthogonal, can be addressed by using bandit models. 
While single-user problem has been investigated intensively so far, multi-user scenario 
is still under investigation. As described in Section \ref{sec:MPMAB}, in a multi-user 
setting, it is important to address the conflict among users by converging to equilibrium 
in some sense. This can be done by using multi-agent bandit algorithms, described before. 
Also, a key point is that the selected channel of users might be sometimes observable. 
Such information can be used as side information to build contextual bandit models. See 
\cite{Maghsudi13:CSNAD} for an example.
\subsubsection{Opportunistic Spectrum Access}
\label{subsec:sensing}
Opportunistic spectrum access in cognitive radio networks can be formulated as a (rested 
or restless) Bernoulli bandit problem. In doing so, each channel is an arm that is either 
in state 0, interpreted as busy, with probability $p$, or in state 1, interpreted as idle, 
with probability $1-p$. Since a secondary user is only allowed to transmit through idle 
channels, a reward is solely associated with state 1. Channel states, as well as parameter 
$p$ are unknown, and a secondary user, modeled as agent, observes the state of the selected 
channel only. Through transmission time, a secondary user aims at maximizing its reward by 
selecting some channel that is available with high probability. The problem can be then 
solved by using indexing policies. An example can be found in \cite{Liu13:LCW}. 
\subsubsection{Sensor Scheduling}
\label{subsec:scheduling}
In sensor networks, the number of sites to be surveyed is very often larger than 
that of available sensors. For instance, for military surveillance and in smart 
grids, it is important to find the places where there is attack or failure. Such 
problems can be addressed by using MABs, where sites are modeled as arms with two 
possible states, similar to the opportunistic spectrum access problem. Furthermore, 
if sensors are used for transmission, and the set of residual energy of each sensor 
is defined on a discrete space, sensor scheduling can be formulated as a MAB problem, 
aiming at minimizing the energy consumption, while maximizing the transmission 
performance. Reference \cite{Chen06:TSSN} is an example. 
\subsubsection{Transmission Mode/Relay Selection}
\label{subsec:relaySelection}
A relay selection problem arises in two-hop (or multi-hop) transmission, where 
multiple relays are available to be used in order to improve the transmission 
performance. To formulate the relay selection problem as a MAB, the reward of 
each relay, modeled as an arm, is defined as the achievable transmission rate 
through that relay, which is the minimum achievable rate of first and second 
hops. Similarly, a problem of transmission mode selection is a natural challenge 
in distributed hierarchical networks, for example, in a D2D communications system 
underlaying a cellular infrastructure. In such a hierarchy, any pair (or group) 
of users can choose to use the traditional cellular transmission mode via a BS, 
or to establish direct communication link. This problem can be simply cast and 
solved by a two-armed bandit model, as proposed in \cite{Maghsudi14:TMSNA}.
\subsubsection{Power Control}
\label{subsec:power}
Power control is one of the less-explored applications of MA-MABs. In order to model 
such problem as a bandit game, a finite discrete set of power levels is considered as 
the set of arms, and the reward process is defined to be some function of SINR, where, 
as in channel selection, interference represents the mutual impact of agents. In comparison 
to channel selection, in a power control problem, the assumption of full or partial 
monitoring is not realistic. That is, agents might not be able to observe the actions, 
i.e., the transmission power levels, of each other. Hence, algorithms should cope with 
the so-called \textit{no monitoring}, where agents' actions are regarded as private. An 
example can be found in \cite{Maghsudi13:JCSPC}.
\subsubsection{Energy Harvesting}
\label{subsec:energy}
Energy harvesting is a relatively new concept attracting increasing interest. In a 
network where nodes are equipped with energy harvesting units, electricity is generated 
from surrounding environment (e.g., solar energy, ambient RF signals). Implementing 
this concept gives rise to some challenges, in particular with respect to scheduling, 
as the energy arrival is in general not deterministic. Under the assumption that the 
energy harvesting process is Markovian, MAB models can be used to address the scheduling 
problem. In the model, similar to the cases before, the reward function is the achievable 
transmission rate while the unknown variant is the energy harvesting process and/or the 
battery state. See \cite{Blasco15:MAC} as an example. Application of MAB theory to this 
area is relatively new and a variety of open problems exist.
\subsubsection{Few Other Applications}
\label{subsec:other}
Although in this article we restrict our attention to wireless resource 
management, it should be noted that MAB models find applications in a variety of other 
networking problems. An important application is \textit{routing}, where on the network 
graph, every end-to-end path consists of multiple edges, and the cost of each edge, 
sometimes time-varying, is unknown a priori. Moreover, after transmitting through each 
path, the only feedback is assumed to be the end-to-end cost of the selected path. Such 
a problem can be modeled and solved using various MAB settings in accordance with the 
network characteristics. See \cite{Awerbuch08:OLOP} for an example. Another field of 
research benefiting from bandit theory is \textit{security}. One example is to combat 
jamming attacks, where an adversary transmits signals to interfere with normal communications 
and temporarily disables the network. Under the assumption that the transmitter and 
the receiver do not pre-share any secrets with each other, the channel selection problem 
can be modeled and solved by using MAB. An example can be found in \cite{Wang12:TOA}.
\subsection{Future Research Directions}
\label{sec:future}
As described in the previous section, a variety of resource management problems have 
been investigated by using bandit models; nonetheless, in most research studies, the 
development of a model and solution is focused on using some infinite-horizon variants 
of bandit problems, with an emphasis on the type of reward evolution, namely, Markovian, 
stochastic or adversarial. In other words, models are often distinguished only based 
on arms' reward generating processes, disregarding many other important and practical 
aspects such as budget constraints, information availability, number of arms, arms' 
dependencies and number of arms to be selected at each round, to name just a few. To 
clarify this shortcoming, some  examples are provided in the following.
\subsubsection{Multimedia Transmission}
\label{subsec:multi}
Most current bandit-theoretical models for wireless networking problems are analyzed 
asymptotically, assuming that the transmission may continue for infinite time. Thereby, 
optimality is defined in terms of long-run discounted reward maximization and/or average 
regret minimization. There are, however, many scenarios such as multimedia transmission, 
where transmission has to be completed under strict delay and/or energy constraints, 
which restricts the number of possible transmission rounds. In such scenarios, algorithms 
that achieve optimality in some long-run sense are not applicable; indeed, here the goal 
is merely to perform some task successfully under some budget constraint rather than 
optimizing the asymptotic performance. Thus, it is imperative to develop new optimality 
conditions, as well as new solution approaches, for applied bandit models in finite horizon. 
\subsubsection{Wireless Sensor Networks}
\label{subsec:WSN}
To the best of our knowledge, current research studies, which apply bandit models 
to solve networking problems, assume that arms, representing resources such as channels or relays, 
are available infinitely often during networking tasks such as channel 
selection or routing. This assumption, however, is not always valid. For instance, in a sensor 
network, a sensor might become unavailable as soon as the energy supply is exhausted. In cognitive 
radio networks, as another example, a channel becomes busy (unavailable) when a primary user starts 
to transmit through that channel. In a bandit model, this effect corresponds to restricted arm 
availability. In other words, each arm might be available only for some limited number of trials. 
Therefore, new definitions of regret and new algorithms need to be developed.
\subsubsection{Correlated Resource Selection}
\label{subsec:Corr}
Another practical issue, currently being neglected, is arms' dependencies. While 
in a great majority of models it is assumed that arms are independent, it is not always the case 
in practice. Indeed, resources, such as relays and channels, might be correlated in many cases. 
For instance, in a cognitive radio with fixed number of primary users, primary channels can be 
considered as dependent since a channel being mostly occupied implies that other channels might 
be less busy. Exploiting such dependencies as a source of information about arms yields efficient 
algorithmic solutions to the bandit, and thereby, the resource management problem. 
\subsubsection{Multi-Agent Setting}
\label{subsec:MAgent}
Besides model-related inefficiencies to be alleviated, there are some open problems in particular 
in multi-agent settings. Although in such settings achieving an equilibrium is 
beneficial for the entire network, there exist a variety of anti-equilibrium reasonings that should 
be addressed. For instance, a general problem is to cope with slow convergence to equilibrium, in 
particular for large number of arms and/or players. Another important problem is equilibrium selection. 
The problem is concerned with guiding the learning agents to the most efficient equilibrium, when 
multiple equilibria exist. In cases, where such convergence is not feasible, an analysis of equilibrium 
efficiency is highly desired. Moreover, in some scenarios, it is desired to have a fair solution 
rather than an efficient one. Note that the problems mentioned above exist also in full-information 
case, but become aggravated in the bandit setting, where only the reward of the activated arm is revealed 
to the player.
%


\section{Efficient 5G small cells with combinatorial (multi-play) bandits}
\label{sec:Example}
In Section \ref{Sec:App}, we briefly studied state-of-the-art, 
thereby clarifying the application of MABs to solve a variety of problems that 
arise in wireless networks. Moreover, we discussed some issues and open problems. 
In this section, we introduce a new application of MABs in next generation wireless 
networks, to design energy-efficient 5G small cells. Primarily, we consider a 
single-agent model with independent arms. In contrast to state-of-the-art, however, 
in our bandit model, multiple arms are selected at each round. We afterwards provide 
some sketch to generalize the model to a multi-agent scenario, and also to the case, 
where arms are dependent, i.e., they affect each other in some specific sense.

Facing an ever-increasing influx in data traffic, 5G networks are foreseen to alleviate this 
problem by deploying dense small cells to underlay the legacy macorcellular networks. 
This takes advantage from low-power and short-range base stations that operate (preferably)  
using the same radio spectrum as the macro base stations and offload macro cell traffic~\cite{hossain2014evolution}. 
In order to maintain low cost, small cells are desired to be self-organized and 
energy-efficient. As a result, a small cell is activated only if it improves 
the overall network performance. Efficient small cell activation can be performed through dynamic small cell 
on/off by a macro cell. Making smart decisions, however, requires some information; 
for instance, the available energy or the number of potential users at each small 
cell, which are both random in nature. Such information is however notoriously difficult 
to acquire specifically when numerous potential small cells exist. Therefore, it 
is reasonable to search for a solution so that a macro cell activates small cells 
in an efficient manner given only limited information. 

Formally, consider a macro BS operating in conjunction with a set $\mathcal{M}$ 
of $M$ potential small cells. Every small cell $m \in \mathcal{M}$ acquires its 
required energy (at least partially) through energy harvesting. That is, while 
sleeping, it gains energy from the environment  through energy harvesting units, 
and uses the stored energy to provide services in its active periods. The amount 
of energy that is gained during each sleeping period depends on environmental 
factors (e.g., the weather) and can change adversarially. The amount of energy 
spent during each servicing period depends on the number of users, which can 
change as well due to user mobility.\footnote{Clearly, both harvested energy and 
number of users can be modeled as stochastic (stationary or non-stationary) variables, 
as well.}~Activating a small cell costs energy even if it does not serve any user; 
thus, at each time (or for every time-period), a macro BS aims at activating a set 
$\mathcal{N} \subseteq \mathcal{M}$ of $N$ small cells. Every activated small cell 
should be able to afford the required energy to provide services to a large enough 
number of users, so that the initial fixed activation cost diminishes. Moreover, it 
is desired to select the subset of small cells in the sense that the overall network 
performance is optimized. The available energy and the number of users in small cells 
are however unknown to the macro BS, which complicates the decision making. We formulate 
this problem as an SA-MAB, by modeling small cells as arms. 

We consider a simple model, where any small cell $m \in \mathcal{M}$ requires $\alpha_{m}$ 
energy units to serve a user with data rate $\beta_{m}$. The cost per energy unit is denoted 
by $r_{m}$. For initial activation, $\kappa_{m}$ units of energy is necessary, even if no 
user is served. Let $A_{m,t}$ and $B_{m,t}$ be the available energy and the number of users 
in small cell at time $t$. Define $\gamma_{m,t}= \max\left \{\left \lfloor \frac{A_{m,t}}
{\alpha_{m}} \right \rfloor,B_{m,t} \right \}$. Then the utility (reward) of selecting a 
small cell $m$ at time $t$ is defined as $g_{m,t}=\gamma_{m,t}\beta_{m}-r_{m}(\gamma_{m,t}
\alpha_{m}+\kappa_{m})$. The total utility of selecting some subset $\mathcal{N}$ therefore 
yields $g_{\mathcal{N},t}=\sum_{m \in \mathcal{N}}g_{m,t}$. 
\begin{figure}[t]
\centering
\includegraphics[width=0.39\textwidth]{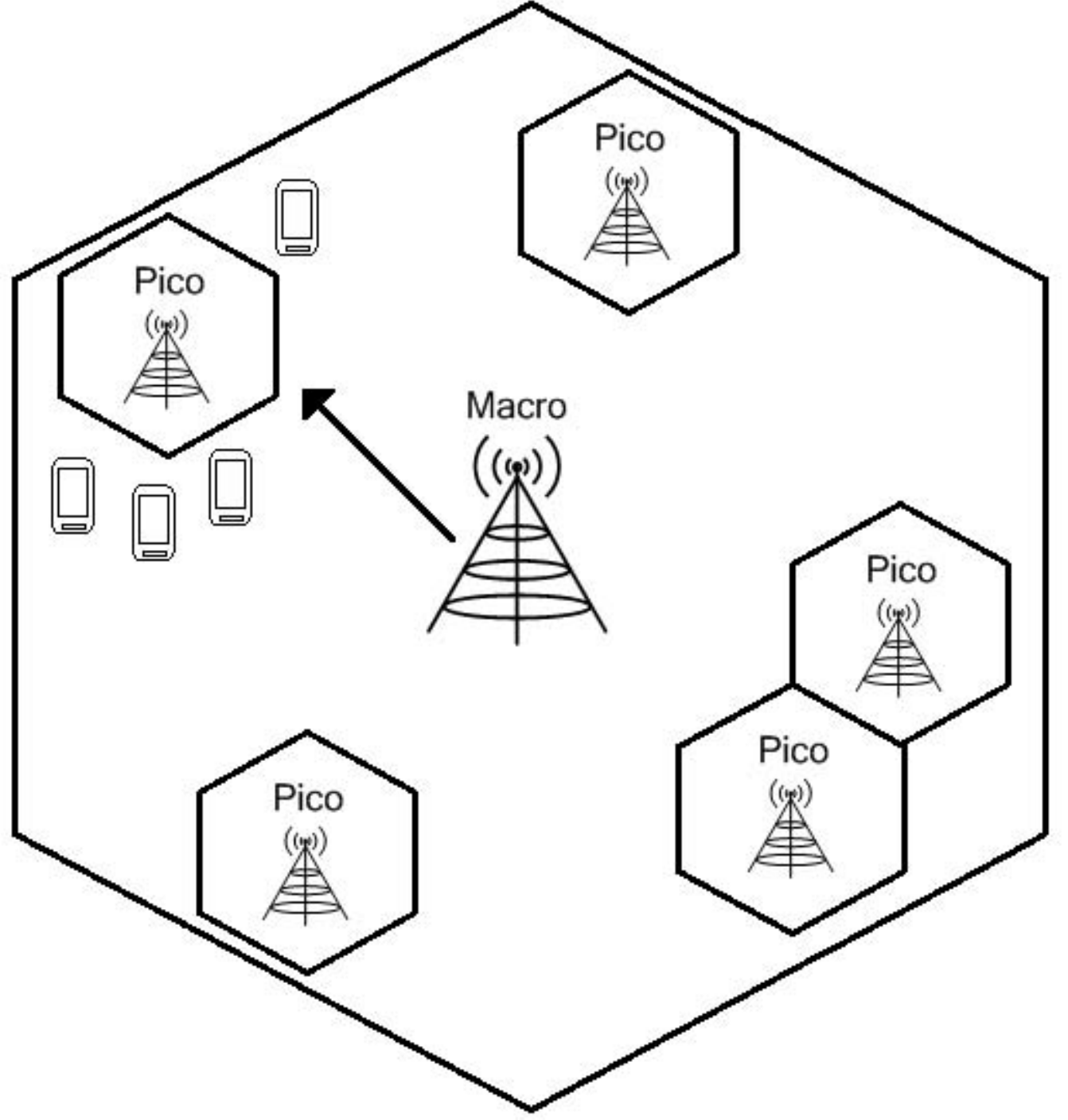}
\caption{Efficient activation of small cells  by macro BS.}
\label{fig:SmallCell}
\end{figure}
Despite similarities, this problem is not identical to the adversarial bandit problem 
described in Section \ref{subsec:adversarial}, due to the following reason. Unlike the 
standard adversarial bandit problem, here multiple arms, and not a single arm, are selected 
at each round of decision making. This type of bandit problems is referred to as \textit{combinatorial} 
or \textit{multiple play} bandits, which stands in contrast to the standard \textit{single 
play} problems. For combinatorial bandits, the conventional definition of regret given 
by (\ref{eq:External}) does not hold. In fact, in order to calculate the regret where 
a subset of arms is selected at each round, the reward achieved by the selected subset 
should be compared with the best \textit{subset} of arms in terms of average reward. The 
problem can therefore be solved by using algorithms described in Section \ref{subsec:adversarial}, 
for instance, weighted-average strategy, by considering each set of $N$ actions as a super 
action, and some other simple adaptations; Nevertheless, such approach is computationally 
inefficient, and therefore, new algorithms are developed that are specifically tailored for 
combinatorial bandit problems. An examples, \textit{EXP3.M}, can be found 
in \cite{Uchiya10:AAB}. It should be however mentioned that most of algorithms are based 
on similar concepts and hence exhibit similar performances.   

\begin{figure}[ht]
\centering
\subfigure[$M=4, N=2$]{
\includegraphics[width=0.40\textwidth]{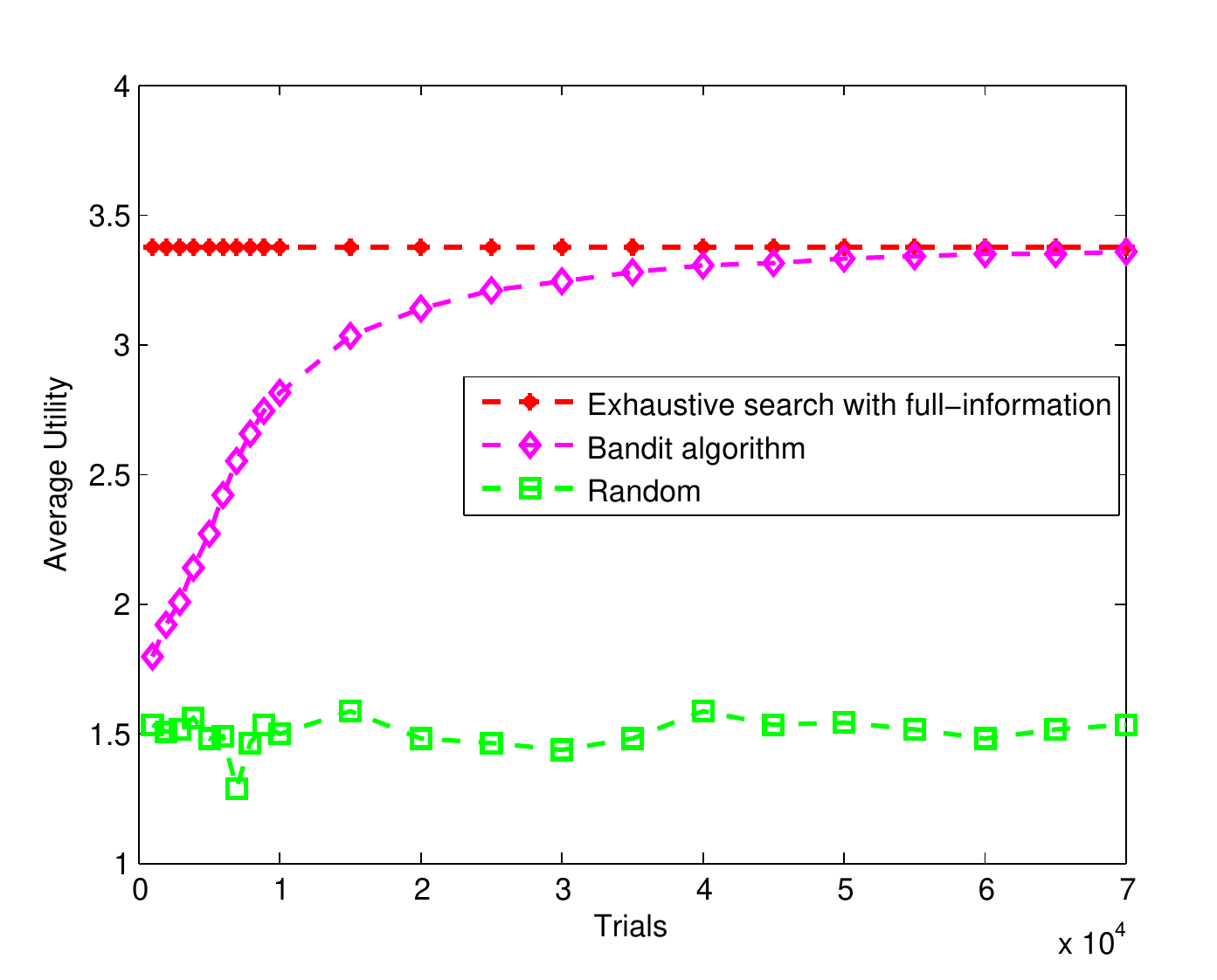}
\label{Fig:Reward42}}
\quad
\subfigure[$M=6, N=3$]{
\includegraphics[width=0.40\textwidth]{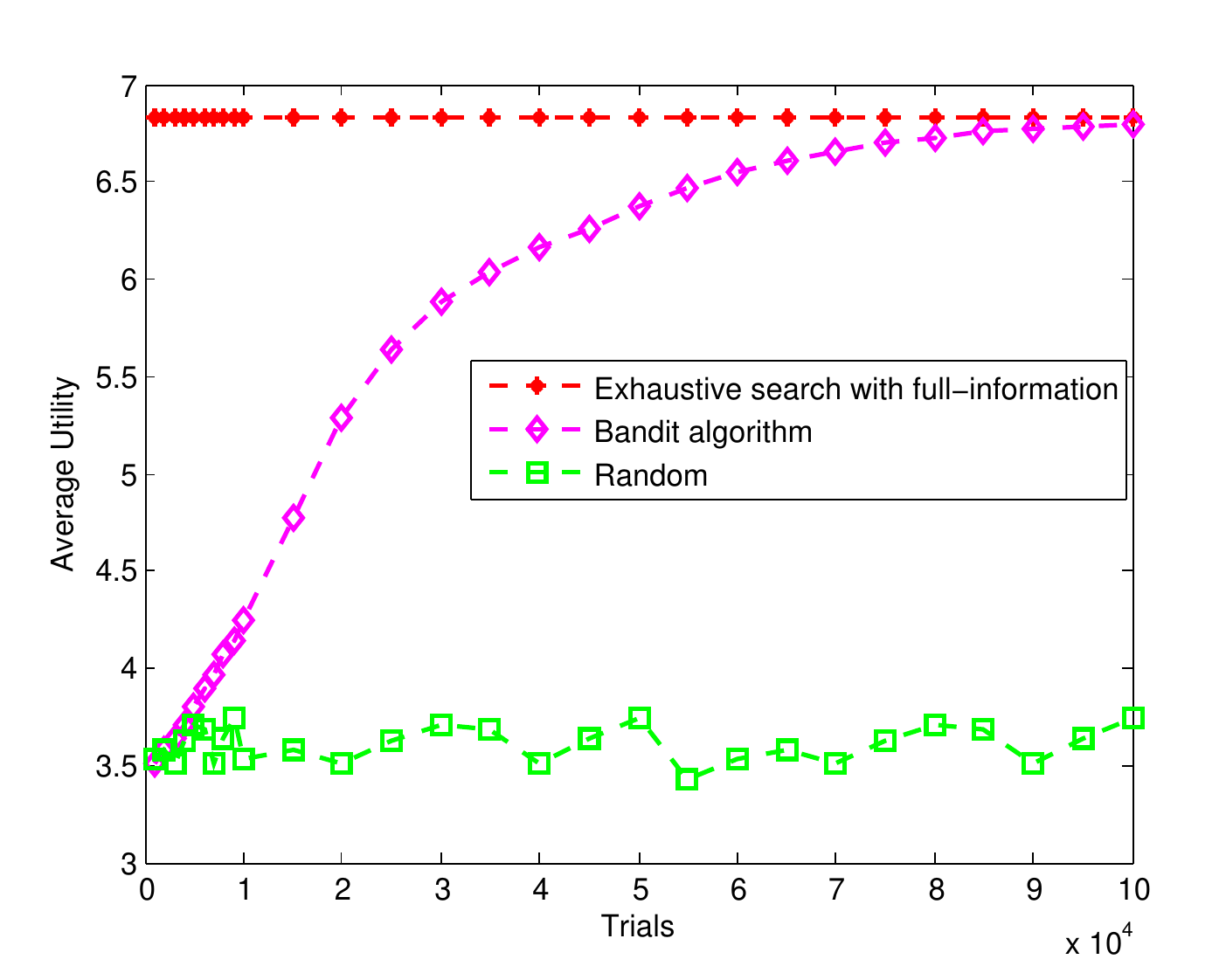}
\label{Fig:Reward63}}
\quad
\subfigure[$M=8, N=4$]{
\includegraphics[width=0.40\textwidth]{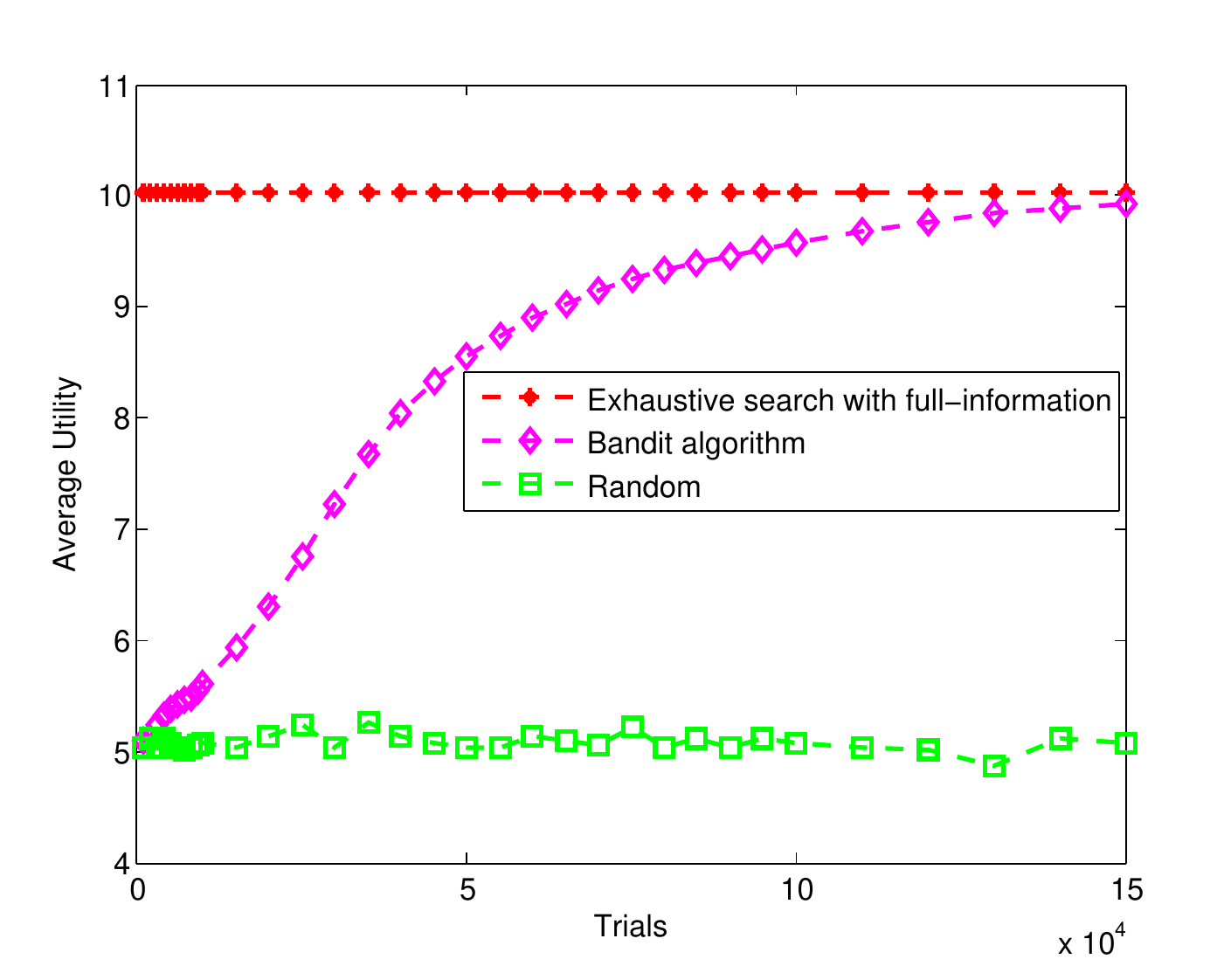}
\label{Fig:Reward84}}
\caption{Average utility achieved by bandit model versus full-information exhaustive search 
for networks of different scales.}
\label{Fig:AvgUti}
\end{figure}

\begin{figure}[t]
\centering
\includegraphics[width=0.40\textwidth]{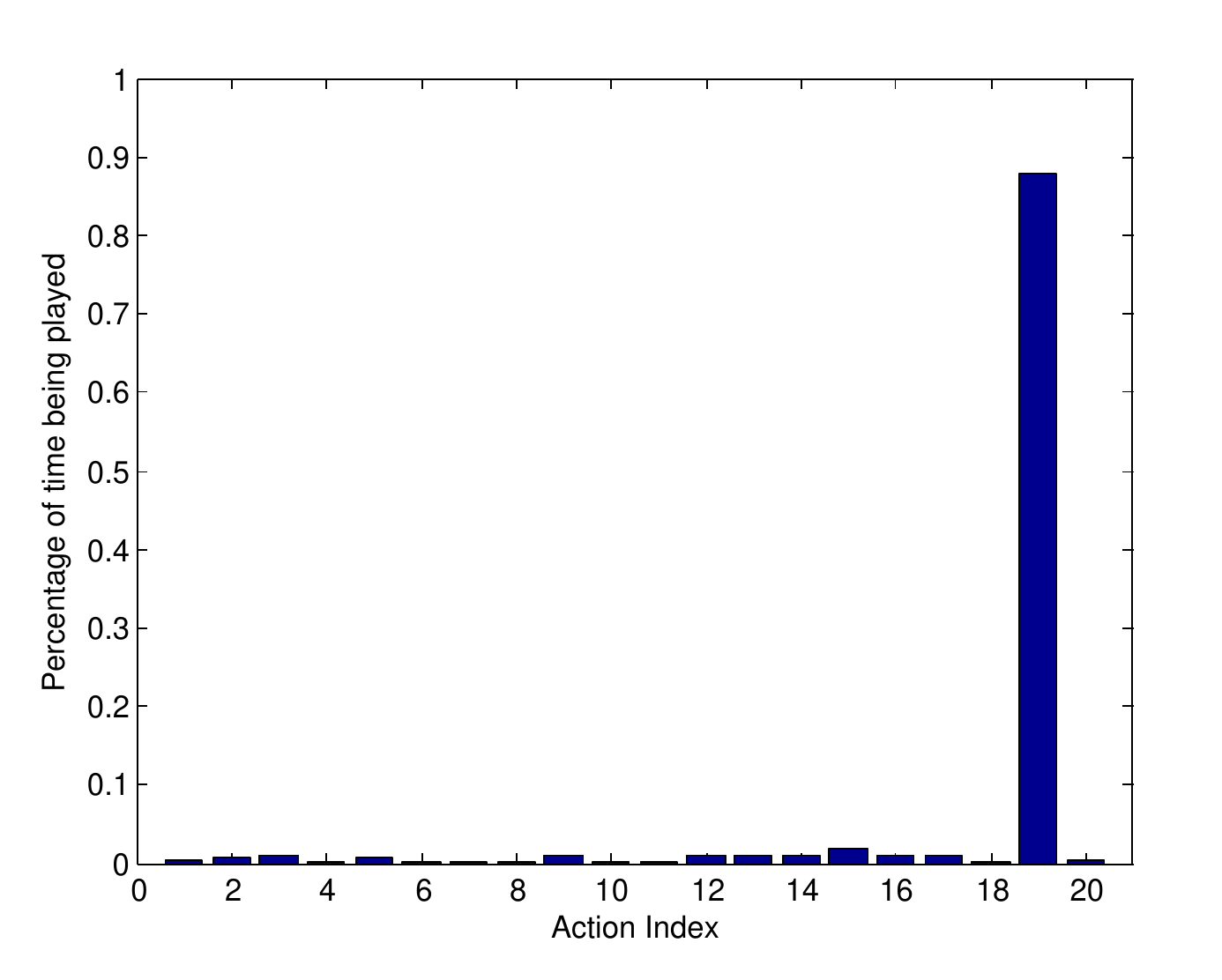}
\caption{Percentage of time each action is played for $M=6, N=3$.}
\label{Fig:Action}
\end{figure}

In order to briefly evaluate the proposed model, we consider three networks with $M=8,6,4$ 
small cells, from which a macro BS selects $N=4,3,2$ to activate, correspondingly. Note that, 
with $M=8$ and $N=4$ for example, there are 60 different 4-tuples of small cells that can 
be selected by the macro BS, which we call \textit{action} and label as $1,...,60$. Number 
of users and residual energy of each small cell is selected randomly \textit{at each trial}, 
without assuming any specific density function. Moreover, the macro BS does not have any 
prior information. To perform selection, the macro BS uses the bandit model described before. 
The average utility versus the best possible subset (in an average sense), which is selected 
through exhaustive search, is shown in Fig. \ref{Fig:AvgUti}. It can be seen that given 
enough time, the utility of the proposed model converges to that of the best selection.
%
For $M=6$ and $N=3$, Fig. \ref{Fig:Action} shows the percentage of time, in which each one 
of the 3-tuples is selected by the macro BS, in $T=5\times10^{5}$ trials. From the figure, 
the best action (here action 19) is played almost all the time. Figures for the other two 
settings are similar, i.e., the best action is played very often. A large fraction of the 
time that is spent to play other actions is the exploration time. In general, the number 
of trials dedicated to exploration depends on the algorithm; nonetheless, in most algorithms, 
exploration time is mostly determined by an exploration parameter, say $\gamma$. Roughly speaking, 
a trial is an exploration trial with probability $\gamma$ and an exploitation trial with probability 
$1-\gamma$. Therefore, a larger value of $\gamma$ yields a larger exploration time on average.

The performance evaluation might be improved for instance by investigating the amount of 
saved energy (cost) in comparison with number of users not being served by small cells, 
 both of which depend on the number of activated small cells, $N$. Thereby, $N$ can be regarded 
as a variable to be optimized according to system's statistical characteristics, instead 
of being fixed and predefined. The question whether such optimization can be included in 
a bandit model or not is an open problem to be investigated in future works.

\begin{figure}[ht]
\centering
\subfigure[Space Complexity]{
\includegraphics[width=0.40\textwidth]{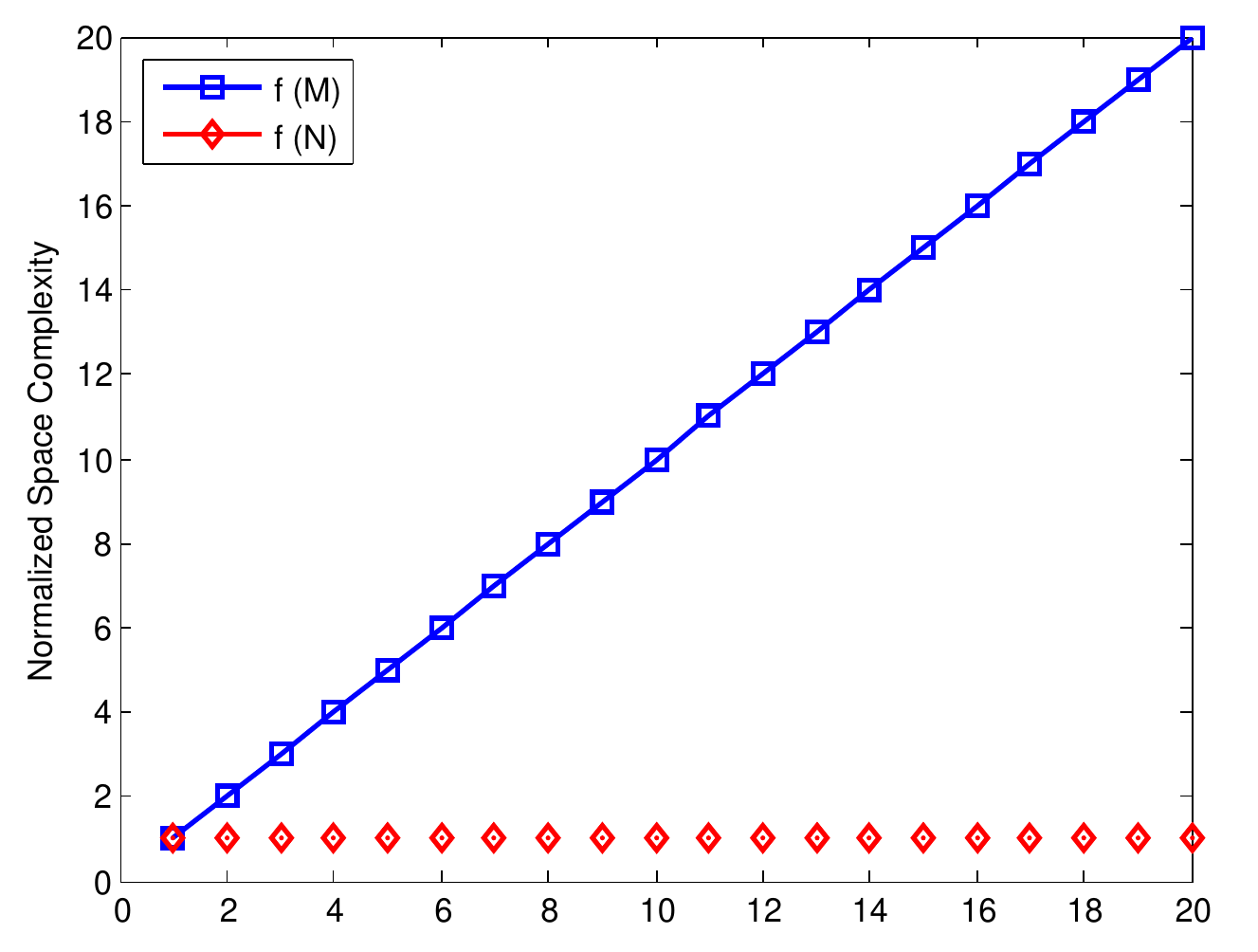}
\label{Fig:SCompl}}
\quad
\subfigure[Time Complexity]{
\includegraphics[width=0.40\textwidth]{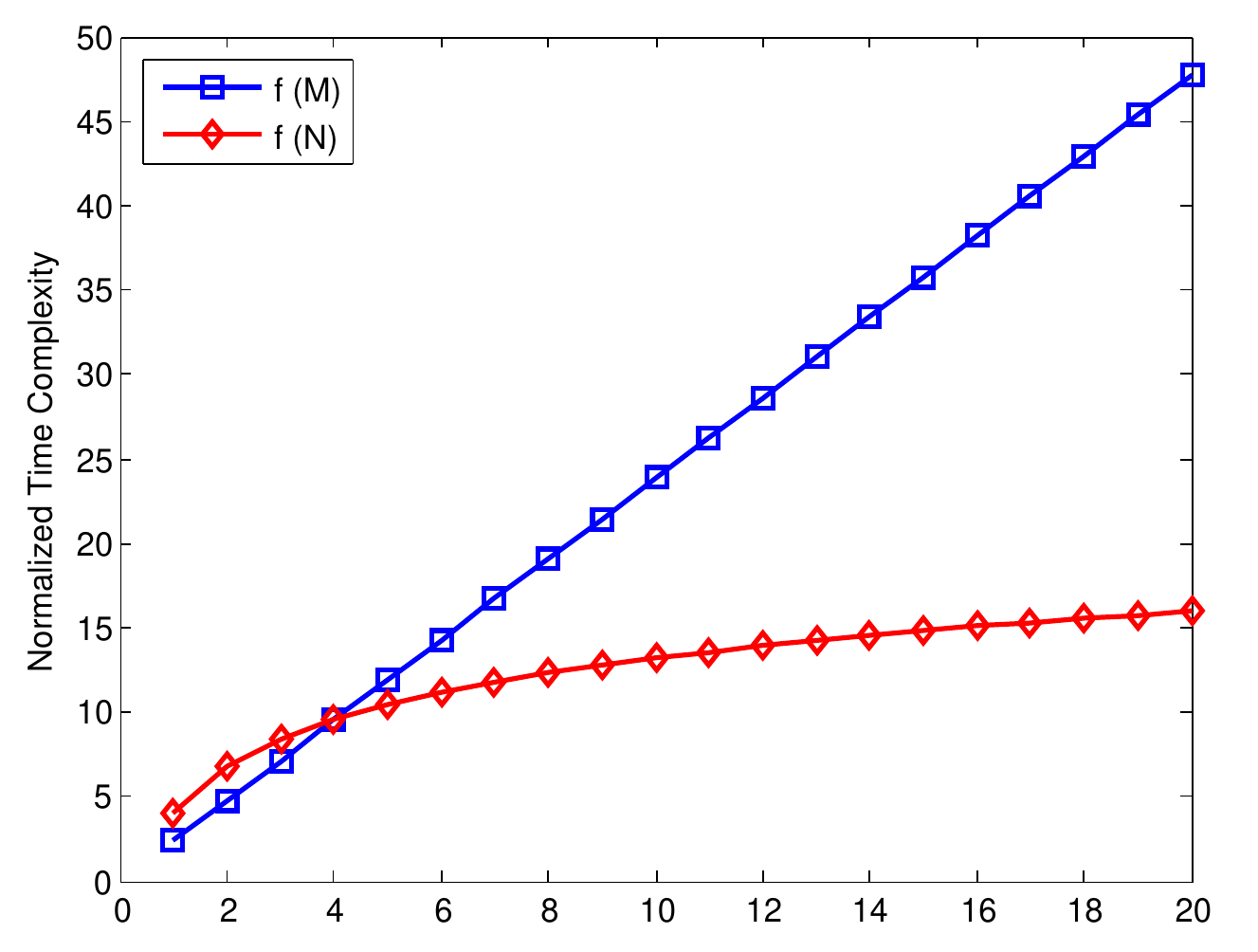}
\label{Fig:TCompl}}
\caption{Scaling space and time complexity of EXP3.M \cite{Uchiya10:AAB} with number of existing 
($M$) and activated ($N$) small cells.}
\label{Fig:Complex}
\end{figure}

It is worth mentioning that the complexity of the proposed small cell planning model depends 
on the algorithm being used to solve the formulated bandit problem. For \textit{EXP3.M} 
\cite{Uchiya10:AAB}, the complexity is of $O(M(\log N+1))$ and $O(M)$, in time and space, 
respectively. 
The growth of complexity  is shown in Fig. \ref{Fig:Complex}. 
Due to linear/logarithmic complexity, the proposed approach can be considered to be scalable with the number of small cells; 
nonetheless, there are scenarios in which distributed solutions to the problem of efficient small 
cell planning are desired. In such scenarios, each small cell decides whether to start the active 
mode or to remain passive. Similar to the centralized case, the decision is based on available energy 
and expected number of users. In case such information is not available at small cells, the problem 
can be modeled as a two-armed bandit problem, with one safe arm (passive mode), and one risky arm 
(active mode). While the passive mode results in no reward, the reward of active mode can be modeled 
as an adversarial or stochastic process, which depends on the energy level as well as number of users 
arriving in each small cell. The problem might have different solutions depending on the type of 
interactions among small cells, which can be competitive or cooperative, for instance. In the event 
small cells are fully decoupled, the problem reduces to $M$ independent two-arm bandit problems. A 
complete investigation of distributed small cell on/off is out of the scope of this article and left 
for our future work.

In addition to considering a distributed model, another direction to improve the current 
model is to take the dependencies of small cells into account. In the formulated problem, 
such dependencies can mainly be concluded based on geographical reasons. In fact, nearby 
small cells are expected to experience similar energy harvesting opportunities (for example, 
during sunny weather in a neighborhood), as well as similar user arrival statistics. Therefore, 
by selecting each small cell and observing the reward, some information can be also derived 
about its neighboring cells. This information is beneficial to shorten the exploration period, 
thereby increasing the accumulated reward or achieving optimal average reward in a shorter 
time. Detailed study of such dependencies is left for our future work.
 
As the last remark, it should be mentioned that the application of MAB models to 5G small 
cell goes beyond the efficient small cell activation problem studied here. An instance is 
the user association problem. While the existence of small cells improves the coverage and 
capacity performances, user association becomes more challenging due to following reasons. 
In sharp contrast to conventional cellular networks, small cells are not always active. In 
addition, the size of each small cell is subject to change as a function of user density, 
for instance. Moreover, a user may associate to a macro BS or a nearby small cell or even 
to both a macro BS and a small cell (or even to multiple small cells). The user association 
problem in a small cell network can be also modeled as a bandit game, either in single-agent 
(centralized) or multi-agent (distributed) settings. While a centralized approach improves 
the possibility of interference coordination, a distributed approach reduces the complexity 
and overhead. Another problem that can be approached by using MAB models 
is inter-cell interference coordination problem in small cell 
networks, which arises due to a dense cell deployment of small cells.


\section{Summary and Conclusion}
\label{Sec:Con}
MAB is a class of sequential decision making problems under strictly limited prior 
information as well as feedback. By providing an overview of MABs, we have argued 
that a wide range of wireless networking problems, including resource management, 
security, routing, scheduling and energy harvesting, can be formulated and solved 
as a bandit problem. We have also reviewed state-of-the-art and applications of MABs, 
with an emphasis on wireless resource allocation. There are a number of open research 
directions to be explored, which have been briefly outlined. Finally, we have provided 
a detailed example of an application of MAB in energy-efficient 5G small cells, where 
the problem of optimal small cell planning is cast as a multi-play (combinatorial) 
bandit problem. Preliminary performance evaluation results have established the 
effectiveness of the proposed model and approach.
%
\bibliographystyle{IEEEbib}
\bibliography{Main}
\end{document}